\documentclass[journal]{IEEEtran}

\usepackage{times}  
\usepackage{helvet}  
\usepackage{courier}  
\usepackage{url}  
\usepackage{graphicx}  
\usepackage{framed,multirow}
\usepackage{amsmath}
\usepackage{mathtools}
\usepackage{bbm}
\usepackage{threeparttable}

\usepackage{amssymb}
\usepackage{latexsym}

\usepackage{url}
\usepackage{xcolor}

\usepackage{hyperref}

\usepackage{graphicx}
\usepackage{tabularx}
\usepackage{subfigure}

\usepackage{diagbox}

\usepackage{type1cm}
\usepackage{lettrine}

\begin{document}
\title{Deep Mining External Imperfect Data for Chest X-ray Disease Screening}
\author{Luyang~Luo,~\IEEEmembership{Student Member, IEEE,}
        Lequan~Yu,~\IEEEmembership{Member, IEEE,}
        Hao~Chen,~\IEEEmembership{Member, IEEE,}
        Quande~Liu,~\IEEEmembership{Student Member, IEEE,}
        Xi~Wang,
        Jiaqi~Xu,
        and~Pheng-Ann~Heng,~\IEEEmembership{Senior Member, IEEE}

\thanks{The work described in the paper was supported in parts by the following grants from Hong Kong Innovation and Technology Fund (Project No. ITS/311/18FP) and National Natural Science Foundation of China (Project No. U1813204).}
\thanks{L.~Luo, H.~Chen, Q.~Liu, X.~Wang, J.~Xu, and P.A.~Heng are with the
Department of Computer Science and Engineering, The Chinese University of
Hong Kong, Hong Kong, China. (e-mails: \{lyluo, hchen, qdliu, xiwang, jqxu, pheng\}@cse.cuhk.edu.hk).}
\thanks{L. Yu is with Department of Radiation Oncology, Stanford University, Stanford, CA 94305, USA. (email:lequany@stanford.edu)}
\thanks{L.~Luo and L.~Yu contribute equally.}
\thanks{H.~Chen is the corresponding author.}
}

\markboth{IEEE TRANSACTIONS ON MEDICAL IMAGING}%
{Shell \MakeLowercase{\textit{et al.}}: Bare Demo of IEEEtran.cls for IEEE Journals}
\maketitle
\begin{abstract}
Deep learning approaches have demonstrated remarkable progress in automatic Chest X-ray analysis. The data-driven feature of deep models requires training data to cover a large distribution. Therefore, it is substantial to integrate knowledge from multiple datasets, especially for medical images. However, learning a disease classification model with extra Chest X-ray (CXR) data is yet challenging. Recent researches have demonstrated that performance bottleneck exists in joint training on different CXR datasets, and few made efforts to address the obstacle. In this paper, we argue that incorporating an external CXR dataset leads to imperfect training data, which raises the challenges. Specifically, the imperfect data is in two folds: \emph{domain discrepancy}, as the image appearances vary across datasets; and \emph{label discrepancy}, as different datasets are partially labeled. To this end, we formulate the multi-label thoracic disease classification problem as weighted independent binary tasks according to the categories. For common categories shared across domains, we adopt task-specific adversarial training to alleviate the feature differences. For categories existing in a single dataset, we present uncertainty-aware temporal ensembling of model predictions to mine the information from the missing labels further. In this way, our framework simultaneously models and tackles the domain and label discrepancies, enabling superior knowledge mining ability. We conduct extensive experiments on three datasets with more than 360,000 Chest X-ray images. Our method outperforms other competing models and sets state-of-the-art performance on the official NIH test set with 0.8349 AUC, demonstrating its effectiveness of utilizing the external dataset to improve the internal classification.
\end{abstract}


\section{Introduction}

\begin{figure*}[htbp]
\centering
\subfigure[Images sampled from NIH and CheXpert.]{
\begin{minipage}[t]{0.48\linewidth}
\centering
\includegraphics[width=3.4in]{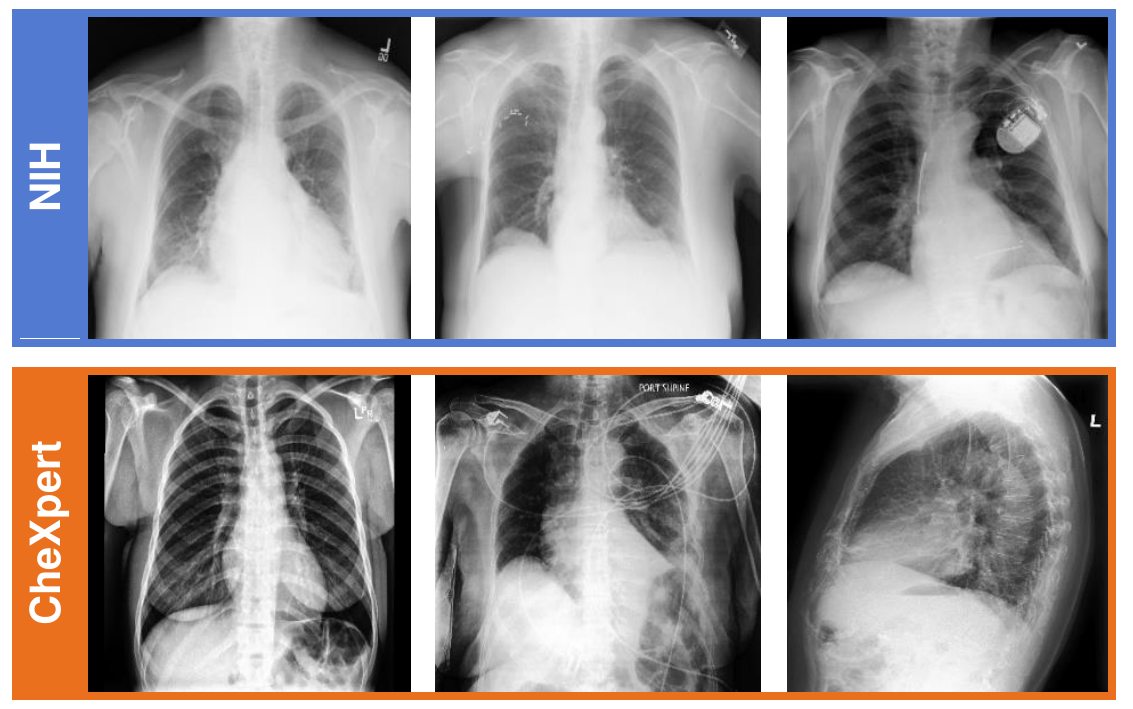}
\end{minipage}
}
\subfigure[Label Distribution of NIH and CheXpert.]{
\begin{minipage}[t]{0.48\linewidth}
\centering
\includegraphics[width=3.4in]{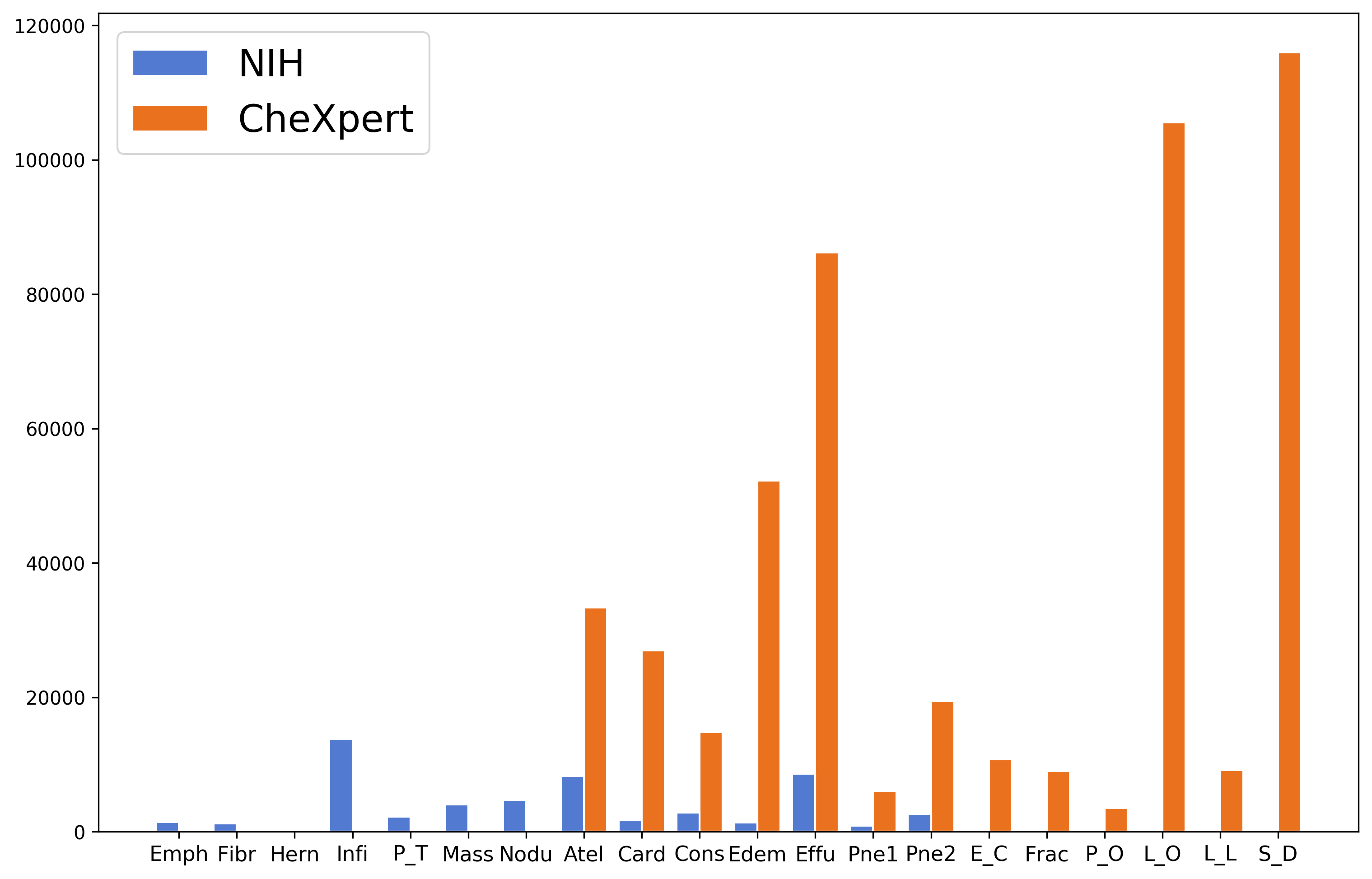}
\end{minipage}
}
\centering
\caption{Comparison between images and label distribution from NIH and CheXpert. Note the CheXpert dataset not only differs from NIH dataset in pixel-wise appearances but also includes more views (about 35\% images in CheXpert are lateral views). Right: Histogram of the label distribution of NIH and CheXpert training sets. CheXpert contains much more common pathologies, as well as many other positive findings. The 20 findings for two datasets are Emphysema (Emph), Fibrosis (Fibr), Hernia (Hern), Infiltration (Infi),  Pleural Thickening (P\_T), Mass, Nodule (Nodu), Atelectasis (Atel), Cardiomegaly (Card), Consolidation (Cons), Edema (Edem), Effusion (Effu),  Pneumonia (Pne1), Pneumothorax (Pne2), Enlarged Cardiomediastinum (E\_C), Fracture (Frac), Pleural Other (P\_O), Lung Opacity (L\_O), Lung Lesion (L\_L), and Support Devices (S\_D). The specific number of each category can be found in Table \ref{Datasets Comparison Table}.}
\label{dataset comparison}
\end{figure*}

\lettrine{C}{HEST} X-ray (CXR) is the most common medical imaging technology in the world, which plays an essential role in diagnosing many thoracic diseases such as pneumonia, atelectasis, lung nodule, etc. The easy and fast feature of CXR leads to a large number of clinical examinations daily. While manual reading of X-rays is expertise-required, time-consuming, and error-prone, automatic detection of diseases is of great value in assisting diagnosing radiographs.\\
\indent Thus far, large-scale datasets of hundreds of thousands of CXR have been released \cite{PLCO,wang2017chestx,CheXpert,MIMIC-CXR}, and the total amounts of publicly available CXR images have added up to near a million. The boost of data scale paves the way for data-driven automatic thoracic disease analysis approaches such as Deep Learning (DL). Nowadays, various DL-based studies \cite{rajpurkar2017chexnet,chen2019lesion,ghesu2019quantifying,pesce2019learning} have been conducted on different CXR datasets. These works focus on developing deep neural networks that can fit on a single dataset with better performances. On the other hand, the data-driven feature of deep learning requires the training data to cover a large distribution. Hence, deep models tend to perform better when incorporating extra datasets. However, joint training on multiple CXR datasets remains a challenging problem. Previous researches have observed that the improvement was limited when employing CXR images from external datasets \cite{yao2019strong,zhang2019improve,lenga2020continual}. For instance, Yao et al. \cite{yao2019strong} developed a binary deep learning model to identify images with pathological findings from healthy cases by jointly training on five CXR datasets. However, the reported AUROC scores on the NIH test set \cite{wang2017chestx} remain unchanged after involving about $500,000$ more training data.\\
\indent We identify the obstacle is raised in that incorporating external datasets leads to imperfect data. As pointed out by Tajbakhsh et al. \cite{tajbakhsh2020embracing}, deep learning requires large, representative, and high quality annotated data. However, it is not practical for medical images to be collected under the same criteria. Instead, different datasets are obtained through inconsistent sampling, imaging, and labeling standard. Therefore, employing an external CXR dataset suffers from imperfect data in two folds. First, the CXR images from different datasets possess \emph{domain discrepancy}. This is caused by factors that affect the image distribution, such as different imaging protocols, inconsistent pre-processing approaches, multiple scanning views, etc. While data-driven approaches highly depend on training data that properly matches the distribution of the test set, incorporating extra data that is out of the initial distribution could not bring benefits to the internal dataset. Second, different CXR datasets are annotated with various labeling strategies, resulting in \emph{label discrepancy}. Specifically, different released datasets focus on different pathological findings. For instance, the NIH dataset is annotated with 14 diseases, and CheXpert \cite{CheXpert} is labeled with 13 categories, while only seven categories are common for the two sets. Subsequently, when dealing with multiple CXR datasets, the X-ray images are \emph{partially labeled} in that some diseases are unknown per image. Fig. \ref{dataset comparison} shows the comparison of the image appearance and category distribution between NIH and CheXpert.\\
\indent Importantly, the mentioned two discrepancies mutually add to the difficulty of each other. For partially labeled data, the images with unknown labels can be regarded as unlabeled data for certain categories. However, learning from unlabeled data usually requires the images to be sampled from the same distribution as the labeled set. Researches \cite{zhang2019improve,PLOSCrossSection} have observed severe performance decay when directly applying a CNN trained on one CXR domain onto another. On the other hand, label discrepancy also poses a challenge of learning domain-invariant features. Since the label distribution has a large difference, the learned discriminative semantics could be used to distinguish features from different domains easily, and the network could fail to obtain better transferability across domains \cite{chen2019transferability}.\\
\indent To tackle these challenges, we propose to learn from the external dataset by simultaneously conquering image and label discrepancies. Since the label spaces between internal and external datasets differ, we formulate the multi-label classification objective of thoracic disease screening as weighted independent binary tasks for each category. To address the domain discrepancy, we present a task-specific adversarial training strategy. Particularly, for each common category, we employ a specific domain discriminator to distinguish the features from each domain. Meanwhile, the network is required to learn more domain-invariant features that can confuse the discriminators. Hence, the network is able to mine the knowledge in the common categories more effectively. Also, to mine the hidden knowledge from the unknown categories, we present temporal ensembling of the model's prediction as soft targets of the unknown categories. As incorporating an external domain would have a higher chance of out-of-distribution predictions, we leverage an uncertainty mechanism to prevent the model from accumulating errors. To this stage, our deep neural network could simultaneously model and address the domain and label discrepancies. We carry out extensive experiments on three CXR datasets, involving more than 360,000 images from NIH, CheXpert, and a private dataset, ImsightCXR. Our method achieves consistent improvements and outperforms other competing approaches, setting state-of-the-art performance on the NIH dataset with AUC of $0.8349$. The contributions of this work are summarized as follows:
\begin{itemize}
  \item We present a novel multi-label chest X-ray disease screening framework that learns to mine additional knowledge from the external dataset.       
  \item We develop a task-specific adversarial training scheme to alleviate the domain discrepancy among CXR datasets and enable the deep model to mine common knowledge more effectively.
 \item We design an effective uncertainty-aware temporal ensembling mechanism to address the label discrepancy and enable the deep model to mine hidden knowledge from unknown labels.
  \item We conduct extensive experiments on three datasets under different settings to demonstrate the effectiveness of our proposed method, and our method achieves state-of-the-art performance on the official NIH test set by mining knowledge from an external dataset.
\end{itemize}

\begin{figure*}[t]
\centering
\includegraphics[width=\textwidth]{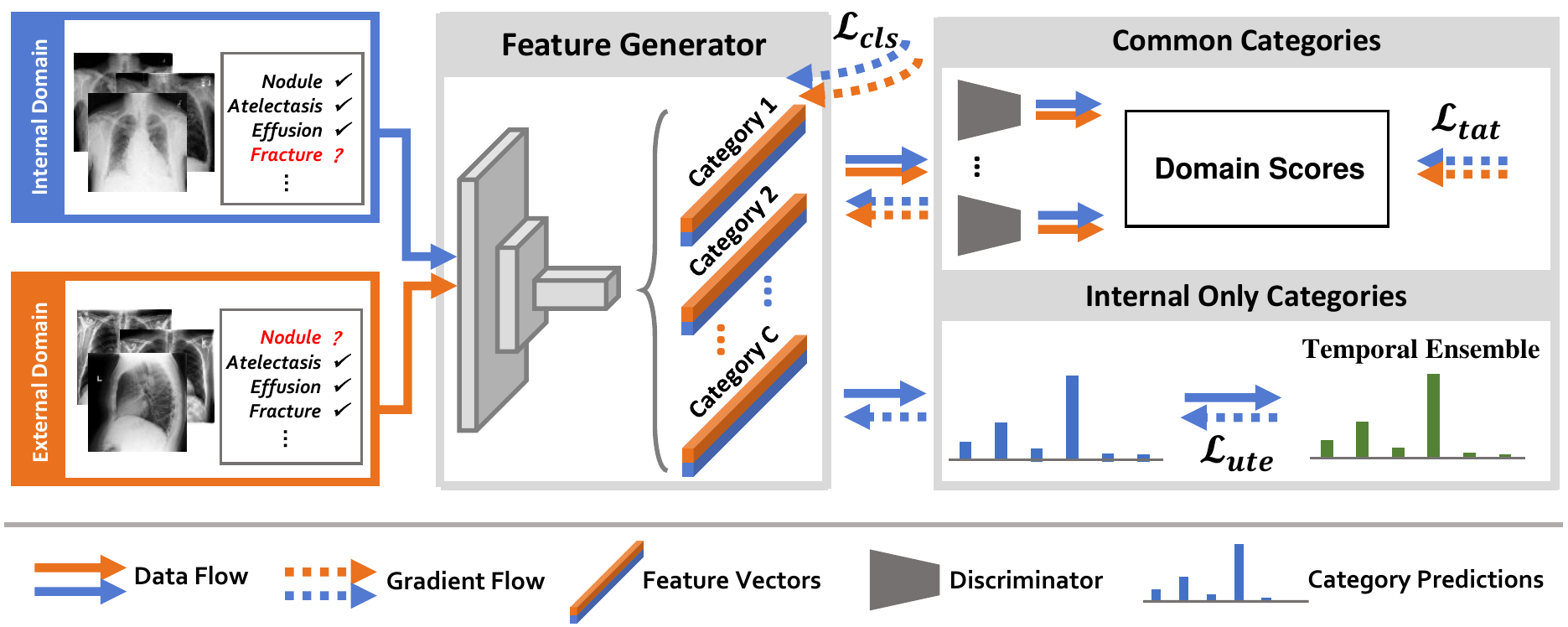}
\caption{Overview of the proposed framework. The inputs are images from two datasets with different appearances and inconsistent label space. The feature generator learns features for each binary classification task and outputs corresponding class scores. The basic objective of the framework is the classification loss. For each common category, a domain discriminator is used to separate features according to the domains. The discriminators are trained with task-specific adversarial loss. For the internal-only categories, the uncertainty-driven temporal ensembling of model predictions is used as soft targets for the missing labels.}

\label{Framework}
\end{figure*}

\section{Related works}
\subsection{Chest X-ray Disease Screening}

Automatic Chest X-ray disease screening has a long history tracing back to the year $1963$ when Lodwick et al. manually designed several descriptors to detect lung cancer \cite{First_CAD_CXR}. It was not until the recent few years did researchers start to take advantage of data-driven approaches, such as Deep Learning, for multi-label classification on large-scale datasets \cite{wang2017chestx,CheXpert}. Wang et al. \cite{wang2017chestx} released the NIH ChestX-ray14 dataset with 112,120 radiographs and conducted a series of benchmark studies with various convolutional neural network (CNN) models. Based on this large dataset, Yao et al. \cite{yao2017learning} adopted a long-short-term-memory network \cite{hochreiter1997long} into a CNN to exploit the dependencies among diseases. Wang et al. \cite{wang2018tienet} further proposed a text-image embedding network to utilize free-text radiological reportscan as prior knowledge to classify and interpret chest X-rays. 
Tang et al. \cite{tang2018attention} adopted a curriculum learning strategy based on weakly-supervised localization of the diseases. To further leverage the lesion location information, Li et al. \cite{li2018thoracic} proposed to use limited bounding-box supervision to improve the classification performance based on multiple instance learning. The most relevant works to ours are from Yao et al. \cite{yao2019strong} and Guendel et al. \cite{guendel2018learning}. The former trained a deep neural network to identify pathological findings from normal radiographs on five datasets and tested it on ten sets. The latter conducted joint training on the NIH dataset and PLCO dataset \cite{PLCO}, and they leveraged the spatial knowledge of PLCO to improve the performance on NIH. However, both works reported limited increases in AUC scores. 

\subsection{Learning from Imperfect External Dataset}
Imperfect data \cite{tajbakhsh2020embracing} could be caused by unlabeled data or partially-labeled data. Recent researches showed that the temporal-ensemble-based approaches could benefit the model by regularizing the predictions on unlabeled data \cite{laine2016temporal,Li2020trans,Liu2020Semi}. Yu et al. \cite{yu2019uncertainty,wang2020ud,xia20203d} further proposed uncertainty-driven strategies that gradually guided the models to learn more robust and stable knowledge. For partially-labeled data, Zhou et al. \cite{zhou2019prior} studied the multi-organ segmentation with partially-labeled domains and developed a prior-aware neural network. To address missing labels in multi-label classification, Yang et al. \cite{yang2016improving} proposed to incorporate structured semantic correlations by utilizing the semantic graph Laplacian as a smooth term in multi-label learning. Wu et al. \cite{wu2018multi} constructed a dependency graph to propagate the label information from given labels to missing labels. Durand et al. \cite{durand2019learning} proposed a curriculum-driven deep convolutional neural network that learns to predict unknown labels. However, all the mentioned works assumed that all datasets were sampled from the same distribution and did not take domain alignment into consideration. \\
\indent To address domain discrepancy among datasets, Rebuffi et al. \cite{rebuffi2017learning} proposed the idea of learning the universal data representations across visual domains. Liu et al. \cite{Liu2020MSNet} further extended universal learning into medical image segmentation. Van et al. \cite{van2018learning} combined different modalities together to enrich the training data. Nevertheless, these methods are often based on fully labeled data, while the label discrepancy among datasets remains unexplored.\\
\indent Another highly related field of medical image analysis dealing with domain discrepancy is domain adaptation. For instance, Wang et al. \cite{wang2019patch,zhang2018task} utilized semantic-aware generative adversarial nets, which are encouraged to generate domain-invariant segmentation mask predictions, to transfer the knowledge learned from one 2D dataset to another domain-discrepant 2D dataset. Zhang et al. \cite{zhang2018translating,chen2020unsupervised} presented unsupervised cross-modality domain adaptation frameworks to adapt convolutional networks trained on one modality to another modality for 3D medical image segmentation. These works usually aimed at adapting models from a labeled source domain to an unlabeled target domain, and the results are usually less efficient than those of the models trained on the labeled target domain.

\section{Methodology}

Fig. \ref{Framework} illustrates the learning framework of our method. In the following section, we first introduce the notations and our weighted partial-label formulation of the imperfect data classification problem. We then present our task-specific adversarial training loss for alleviating the feature level domain difference and mining the common knowledge in common categories. Finally, we introduce the uncertainty-aware temporal ensembling for mining hidden knowledge from unknown categories.

\subsection{Basic Framework for Partial-label Classification}
Let $X^{d} = \{(I^{d_{1}},y^{d_{1}}),...,(I^{d_{n}},y^{d_{n}})\}$ denote the $d^{th}$ dataset involved in our presented method. $I^{d_{i}}$ is the $i^{th}$ CXR image from dataset $X^{d}$, and $y^{d_{i}} = [y^{d_{i}}_{1},...,y^{d_{i}}_{C}] \in Y^{d} \subseteq \{0,1\}^{C}$ is the label vector. Given a sample $i$ from the dataset $X^{d}$ and a category $c$, $y^{d_{i}}_{c} = 1$ (resp. $0$ and $-2$) represents the category is present (resp. absent and unknown). Further, $l^{d}$ denotes the fully labeled target space of dataset $X^{d}$, and $\cap l$ represents the intersection of labeled target spaces.\\
\indent Automatic CXR image screening can be regarded as multi-label classification, i.e., a mapping function from a given image $I$  to a label vector $y \in \mathbb{R}^{C}$. We formulate the objective into a combination of $C$ weighted independent binary classification tasks as follows:
\begin{equation}\label{Weighted BCE}
\begin{split}
    \mathcal{L}_{cls} = -\frac{1}{C}\sum^{C}_{c=1}\alpha_{c}[\beta_{c} \cdot&\mathbbm{1}_{[y_{c}=1]}\log(p_{c})+ \\
    &\mathbbm{1}_{[y_{c}=0]}\log(1-p_{c}) ]
\end{split}
\end{equation}
\noindent where $\mathcal{L}_{cls}$ is the partial-label classification loss, $p \in \mathbb{R}^{C}$ is the prediction vector of a deep neural network, $\mathbbm{1}_{[\cdot]}$ is an indicator function, $\alpha$ and $\beta$ are weighting parameters. Particularly, $\beta_{c} = \frac{N_{c}}{P_{c}}$ with $N_{c}$ the total number of negative samples (with label $0$) and $P_{c}$ the total number of positive samples (with label $1$) of category $c$. Additionally, inspired by \cite{chen2018gradnorm}, we find the independent binary classification tasks also have different difficulties. Consequently, some tasks may dominate the overall gradient of the training process. To this end, we assign different weights for each binary classification task with $\alpha_{c}$ to balance the gradients of different tasks. Specifically, the task weights are set according to whether the classification task is shared across domains. As a result, an AUC improvement is observed when more weights are assigned to the common tasks, which would be shown in later experiments.

\subsection{Task-specific Adversarial Training for Common Knowledge Mining} \label{Task-specific Adversarial Training}

As aforementioned, severe domain bias exists in between CXR datasets, which causes limited or even decayed performance with joint training. To tackle this challenge, we propose to alleviate the feature-level domain discrepancy through adversarial training. Let the multi-label classification neural network be denoted as $\mathcal{F}(\mathcal{G}(I))$ with $\mathcal{G}$ being the feature generator and $\mathcal{F}$ the feature classifier, i.e., the last fully connected layer (FC-layer). The adversarial training can be applied by a mini-max game in between a domain discriminator $\mathcal{D}$ and the feature generator $\mathcal{G}$:
\begin{equation}
\begin{split}
    \mathcal{L}_{adv}(\mathcal{G},\mathcal{D},I) = &\mathbb{E}_{\mathcal{G}(I^{d})\sim X^{d}}[\log \mathcal{D}(\mathcal{G}(I^{d})]
 \\
             &\mathbb{E}_{\mathcal{G}(I^{-d})\sim X^{-d}}[\log(1-\mathcal{D}(\mathcal{G}(I^{-d}))]
\end{split}    
\end{equation}
\noindent where $\mathcal{L}_{adv}$ is the adversarial training loss, $\mathcal{G}(I^{d})$ is the feature learned from domain $d$, $\mathcal{G}(I^{-d})$ is the feature from other domains. The feature generator $\mathcal{G}$ aims to minimize the objective against an adversary $\mathcal{D}$ that tries to maximize it. Hence, the objective above is $\mbox{min}_{\mathcal{G}}\mbox{max}_{\mathcal{D}}\mathcal{L}_{adv}(\mathcal{G},\mathcal{D})$. In this way, the feature generator would learn to generate features that are more domain-invariant. \\
\indent Generally, $\mathcal{G}$ ends with a global average pooling layer, and $f = \mathcal{G}(I) \in \mathbb{R}^{N}$ is a $N$-dimensional feature vector. The FC-layer $\mathcal{F}$ performs a matrix multiplication that maps $f$ into a prediction vector of dimension $C$. We notice such design brings too rich semantics to $f$ under the multi-label scenario, where $2^{C}$ combinations of categories potentially exist. Hence, the label distributions difference could be easy evidence to distinguish the source of the features. To cope with this problem, we append $C$ parallel FC-layers of dimension $N\times N'$ to $\mathcal{G}$. Hence, the feature generated by the new generator $\mathcal{G}'$ is of dimension $N' \times C$, where a vector $f_{c}=\mathcal{G}'(I)_{c}$ captures the semantic of the $c$-th category. Correspondingly, we replace $\mathcal{F}$ with $C$ classifiers of dimension $N' \times 1$ and replace the single discriminator $\mathcal{D}$ with several discriminators specified for each common category. In this way, we extend the mini-max game into several category-specified sub-games as follows:
\begin{equation}
\begin{split}
    \mathcal{L}_{tat}(\mathcal{G}',\mathcal{D'},I) = \sum_{c\in \cap l}^{C} &\mathbb{E}_{\mathcal{G}'(I^{d})_{c}\sim
X^{d}}[\log \mathcal{D}_{c}(\mathcal{G}'(I^{d})_{c})] + \\
                          &\mathbb{E}_{\mathcal{G}'(I^{-d})_{c}\sim X^{-d}}[\log(1-\mathcal{D}_{c}(\mathcal{G}'(I^{-d})_{c}))]
\end{split}    
\end{equation}
\noindent where $\mathcal{D'} = \{\mathcal{D}_c|c\in \cap l\}$. Hence, the objective becomes $\mbox{min}_{\mathcal{G'}}\mbox{max}_{\mathcal{D'}}\mathcal{L}_{tat}(\mathcal{G'},\mathcal{D'})$. Note that we specifically apply the adversarial training on the intersection category set, i.e., $c\in \cap l$. Therefore, mining the common knowledge of the common categories is eased with the specified features being more domain-invariant. Recall that we formulate the multi-label classification into multiple independent binary classification tasks for each category, the above objective is referred to as Task-specific Adversarial Training (TAT) loss.

\subsection{Uncertainty-aware Temporal Ensembling for Hidden Knowledge Mining}
As can be observed in the above framework, the CXR datasets possess label discrepancy raised by partial labels, e.g., Nodule is labeled in the internal domain while unlabeled in the external domain. For such categories with missing labels, the unlabeled data could potentially provide additional information by assigning pseudo-labels to the unlabeled data \cite{durand2019learning}. However, in our case, the external and internal datasets obtain domain discrepancy, and hard labeling (giving labels by either 0 or 1) is hence more prone to accumulation of prediction errors \cite{berthelot2019mixmatch}. Additionally, incorporating hard labels changes the ratio between positive and negative samples, which consequently changes the optimization for the weighted classification loss. \\
\indent To this end, we propose to utilize soft labels as the target for unknown categories. Drawing spirit from the recent advances on semi-supervised learning \cite{laine2016temporal}, we take advantage of the exponential moving average (EMA) of the model prediction to assign pseudo-labels. Denoting the prediction of the deep neural network at epoch $t$ as $p_{t}$, the temporal ensembling is updated as follows:
\begin{gather}
   Z_{t} = \gamma Z_{t-1} + (1-\gamma)p_{t-1}\mbox{      and      } z_{t} = Z_{t} / (1-\gamma^{t-1})\\
   \mathcal{L}_{tem}= \mathbbm{1}_{[y_{c}=-2]}\left\| p_{t}-z_{t} \right\|_{2}^{2}
\end{gather}
\noindent where $Z_{t}$ is the EMA of the prediction, $\gamma$ is a momentum term that controls how far the ensemble reaches into training history, $1-\gamma^{t}$ is a bias correction term in line with previous studies \cite{laine2016temporal,kingma2014adam}, $z_{t} \in [0,1]$ is the final soft target for regularization, and $\mathcal{L}_{tem}$ is the loss supervised by the temporal ensemble. Note at the first epoch, i.e., $t=1$, $Z_{1}$ is set to all zeros. To compute the loss with the temporal ensemble, we leverage the squared $\mathcal{L}_{2}$ norm. Unlike the cross entropy loss, this objective is bounded and less sensitive to incorrect predictions \cite{berthelot2019mixmatch}. Moreover, the EMA procedure ensures that the predictions of later epochs contribute more, which enables the temporal ensemble to be more stable and confident.

\indent With the proposed temporal-ensembling-based labeling strategy, the pseudo-labels are less prone to error accumulation than hard labels. However, as mentioned before, our internal and external datasets have different domain distributions, whereas semi-supervised learning highly relies on the assumption that the labeled and unlabeled data are sampled from the same distribution. For the internal-only categories, the \textit{out-of-distribution} image from the external dataset would lead to considerable \textit{uncertain} predictions. As pointed out in \cite{han2018co}, a deep model would give larger classification losses for the uncertain labels. Since the magnitude of cross-entropy loss depends on how close the prediction is to $0.5$, we propose to filter out the uncertain predictions by the following:
\begin{gather}
   \mathcal{L}_{ute}= \mathbbm{1}_{[y_{c}=-2, |0.5-p_{t}|_{1}\geq H]}\left\| p_{t}-z_{t} \right\|_{2}^{2}
\end{gather}
where $|0.5-p_{t}|$ is the estimated uncertainty of the prediction, and $H$ is a threshold that gradually decreases. At the starting epochs, $H$ is large, and only the predictions with high certainty would be selected in the soft label objective. Hence, we are able to first stabilize the prediction of unknown categories. As the model converges and stabilizes, $H$ decreases, and more predictions are then involved in the computation. With our uncertainty-aware temporal ensembling (UTE) loss, the model is encouraged to learn more reliable knowledge step by step from the data with missing labels. \\
\indent Note that UTE differs from the consistency-based methods \cite{laine2016temporal} that require many data augmentation strategies during training. We mainly take advantage of the stable and robust smooth labels generated by the temporal ensemble scheme. The uncertainty mechanism further encourages the model to learn from confident samples gradually. Hence, our method could perform well even without heavy augmentation, as would be shown in Section \ref{Approaches Incorporating External Dataset}.

\subsection{Network Architecture and Training Scheme}
We leverage DenseIBN-121 \cite{pan2018two} as the backbone network. DenseIBN-121 adopts densely skip connections as same as DensNet-121 \cite{huang2017densely}, while the only difference is that DenseIBN-121 adopts Instance Normalization \cite{ulyanov2016instance} together with Batch Normalization \cite{ioffe2015batch} in the Dense Blocks. It shows better generalization under cross-domain circumstances, as demonstrated in the original paper. We modify the classifier layer to $C$ branches, each with two fully connected layers as described in section \ref{Task-specific Adversarial Training}, and initialize other layers from the pre-trained model on ImageNet \cite{deng2009imagenet}. The discriminators are simply 3 FC-layers with 2 LeakyReLU \cite{maas2013rectifier} layers. The overall objective function is as follows:
\begin{equation}
    \mathcal{L} = \mathcal{L}_{cls}+\lambda_{tat} \mathcal{L}_{tat} + \lambda_{ute} \mathcal{L}_{ute}
\end{equation}
\noindent where $\lambda_{tat}$ and $\lambda_{ute}$ are the weights for adversarial training loss and unsupervised loss, respectively.\\
\indent During training, we load the CXR images with a size of $320\times 320$ as the input. Following \cite{rajpurkar2017chexnet}, we only use horizontal flipping as the data augmentation strategy. The images from different datasets are fed into the network with an equal batch size at each step. We use Adam \cite{kingma2014adam} as the solver to the DenseIBN structure and RMSprop \cite{Tieleman2012} to optimize the discriminators. The learning rate of Adam is initially set to $1e-4$ and decayed to $1e-5$ and $1e-6$ after epoch 3 and 6, while the learning rate for RMSprop optimizer are fixed to $1e-4$. Based on our preliminary experiments, the adversarial loss is directly trained together with other losses in every iteration. The whole framework is implemented with PyTorch \cite{paszke2019pytorch} on one Titan XP GPU. We train the model for 8 epochs until it converges. For the hyperparameters, we set $\alpha_{c \in \cap l}=3$ for common classes and $\alpha_{c \in l^{d}\backslash \cap l}=1$ for other classes. For task-specific adversarial training and uncertainty-aware temporal ensembling, we let $\lambda_{tat}=0.03$, $\lambda_{ute}=30$, and $\gamma=0.9$.

\section{Experiments and Results}
In this section, we first introduce the three datasets involved in our study and the evaluation metrics we employed. We then demonstrate our experimental results on the NIH dataset, where we incorporate CheXpert as the external data. \textcolor{black}{Afterward}, we present the results on our own collected dataset, ImsightCXR, where NIH is used as the external site. \textcolor{black}{In the last sub-section}, we show the ablation study on different components of our method.
\subsection{Datasets}

\begin{table}
\centering
\caption{The Pathological Findings in NIH, CheXpert, and ImsightCXR Datasets. For CheXpert, we report the summation of positive findings and uncertain labels.}
\label{Datasets Comparison Table}
\begin{tabular}{l|r|r|r} 
\hline
Dataset                    & \multicolumn{1}{l|}{NIH} & \multicolumn{1}{l|}{CheXpert} & \multicolumn{1}{l}{ImsightCXR}  \\ 
\hline
Emphysema                  & 2,516                    & -                             & -                            \\
Fibrosis                   & 1,686                    & -                             & 2,557                        \\
Hernia                     & 227                      & -                             & 35                           \\
Infiltration               & 19,871                   & -                             & 178                          \\
Pleural Thickening         & 3,385                    & -                             & 4,443                        \\
Mass                       & 5,746                    & -                             & 1,244                        \\
Nodule                     & 6,323                    & -                             & 4,387                        \\
Atelectasis                & 11,535                   & 58,710                        & 1,099                        \\
Cardiomegaly               & 2,772                    & 29,599                        & 2,594                        \\
Consolidation              & 4,667                    & 36,706                        & 492                          \\
Edema                      & 2,303                    & 60,476                        & -                            \\
Effusion                   & 13,307                   & 85,115                        & 5,275                        \\
Pneumonia                  & 1,353                    & 20,234                        & 5,588                        \\
Pneumothorax               & 5,298                    & 19,976                        & 2,853                        \\
Enlarged Cardiomediastinum & -                        & 19,168                        & -                            \\
Fracture                   & -                        & 7,754                         & 1620                         \\
Pleural Other              & -                        & 4,212                         & -                            \\
Lung Opacity               & -                        & 97,010                        & -                            \\
Lung Lesion                & -                        & 7,927                         & -                            \\
Support Devices            & -                        & 106,729                       & -                            \\
Tuberculosis               & -                        & -                             & 10,009                       \\
\hline
\end{tabular}
\end{table}

\begin{table}[t]
\color{black}
\centering
\caption{AUC results of cross-sectional testing on the common six categories among NIH, CheXpert and ImsightCXR.}
\label{Cross Sectional Testing}
\begin{tabular}{|l|c|c|c|} 
\hline
\backslashbox{Testing}{Training}           & NIH   & CheXpert & ImsightCXR  \\ 
\hline
NIH        & 0.802 & 0.669    & 0.704       \\ 
\hline
CheXpert   & 0.799 & 0.848    & 0.767           \\ 
\hline
ImsightCXR & 0.807 & 0.801        & 0.917       \\
\hline
\end{tabular}
\end{table}

We conduct experiments on three Chest X-ray datasets, including NIH, CheXpert, and ImsightCXR, a private dataset collected from multiple Chinese hospitals. The NIH dataset contains 112,120 frontal-view CXR images from 32,717 patients, where each image is labeled with 14 possible pathological findings (absent or present). The CheXpert dataset obtains 224,316 chest radiographs of 65,240 patients with both frontal and lateral views available. For experiments on these two public sets, we adopt the official split for training, validation, and testing. Hence, \textcolor{black}{a total of 303,646 images} are used for training, from which 80,232 are from the NIH database, and 223,414 are from CheXpert. We use 6,292 images from NIH for validating, and 25,596 images for testing. U-Ones \cite{CheXpert} strategy is used for CheXpert. The ImsightCXR dataset has in total 32,261 frontal CXR images taken from 27,253 patients. Specifically, each image of ImsightCXR is labeled by three radiologists with the corresponding radiological reports. Thus, the annotation of ImsightCXR is much cleaner than the other two \textcolor{black}{that} are annotated with natural language processing. Table \ref{Datasets Comparison Table} reports the exact numbers of positive findings in each dataset. We split ImsightCXR (without overlap of patients) into the training set, validation set, and testing set with images of 13,963, 5,118, and 13,180 images, respectively. 

\textcolor{black}{It is worth noting} that the label "No Finding" in each dataset does not mean the subject is pathologically healthy. For example, CheXpert considers "Emphysema", "Hernia", and "Thickening" as "No Finding", as can be \textcolor{black}{referred to in} the labeler tool of CheXpert \footnote{https://github.com/stanfordmlgroup/chexpert-labeler}. Besides, "Mass" and "Nodule" are regarded as sub-classes of "Lung Lesion", "Infiltration" is a sub-class of "Lung Opacity". CheXpert also takes "Fibrosis" as a sub-class of "Pleural Other". Therefore, we could tell that the unknown categories are ignored by the labeling strategy instead of being necessarily absent.

\subsection{Evaluation Metrics}
Following previous studies \cite{wang2017chestx, wang2018tienet}, we employ the area under the receiver operating characteristic curve (AUC) to evaluate our method and compare it with other approaches. The AUC of each disease is evaluated as well as the average AUC of all categories. To validate the effectiveness of our specific designs, we also compute the averaged AUC of the common categories and averaged AUC of internal-dataset-only classes. Unless specified in the following sections, we inference un-augmented images only once and use the single output for the validation and testing phase. We validate the model every 3600 steps and choose the one with the highest validation AUC for follow-up evaluation.

\subsection{\textcolor{black}{Analysis of Datasets Discrepancy}}

\textcolor{black}{Apart from the label discrepancy clearly showed in Table \ref{Datasets Comparison Table}, we report in Table \ref{Cross Sectional Testing} the cross-sectional testing on the six common categories among the three datasets with the mentioned splits. As can be observed, directly applying a single-dataset-trained model onto another set would lead to severe performance drop, which indicates the distribution of the testing set is quite different from that of the training set. In particular, testing the models trained with the external datasets on the internal datasets (CheXpert to NIH and NIH to ImsightCXR) would cause at least 11 points decaying over the common categories. The experiments here validate the domain discrepancies among the three datasets.}

\begin{table*}[ht]
\caption{Classification Results On NIH Dataset Incorporating CheXpert as External Dataset}\label{External Domain Comparison Table}
\centering
\resizebox{\textwidth}{!}{\begin{tabular}{c|cccccccccccccc|c} 
\hline
\multirow{2}{*}{Method}             & \multicolumn{7}{c|}{NIH Only}                                                                                                                                                                            & \multicolumn{7}{c|}{Common}                                                                                                                                                                               & \multirow{2}{*}{Mean}       \\ 
\cline{2-15}
                                    & Emph                       & Fibr                       & Hern                       & Infi                       & P\_T                        & Mass                       & \multicolumn{1}{c|}{Nodu}  & Atel                       & Card                       & Cons                       & Edem                       & Effu                       & Pne1                       & Pne2                        &                             \\ 
\hline\hline
\multicolumn{1}{c|}{Single Dataset} & \multicolumn{1}{l}{0.9368} & \multicolumn{1}{l}{0.8166} & \multicolumn{1}{l}{0.9358} & \multicolumn{1}{l}{0.7066} & \multicolumn{1}{l}{0.7756} & \multicolumn{1}{l}{0.8156} & \multicolumn{1}{l}{0.7684} & \multicolumn{1}{l}{0.7619} & \multicolumn{1}{l}{0.8936} & \multicolumn{1}{l}{0.7451} & \multicolumn{1}{l}{0.8456} & \multicolumn{1}{l}{0.8272} & \multicolumn{1}{l}{0.7174} & \multicolumn{1}{l|}{0.8669} & \multicolumn{1}{l}{0.8152}  \\
Joint Training                      & 0.9368                     & 0.8199                     & 0.9349                     & 0.7138                     & 0.7918                     & 0.8205                     & 0.7725                     & 0.7590                     & 0.9024                     & 0.7486                     & 0.8296                     & 0.8279                     & 0.7352                     & 0.8870                      & 0.8200                      \\
Multi-Task \cite{guendel2018learning}                          & 0.9344                     & 0.8107                     & 0.9239                     & 0.7120                     & 0.7827                     & 0.8151                     & 0.7698                     & 0.7603                     & 0.8981                     & 0.7449                     & 0.8474                     & 0.8274                     & 0.7207                     & 0.8798                      & 0.8162                      \\ 
DNetLoc \cite{guendel2018learning}                             & 0.8950                     & 0.8180                     & 0.8960                     & 0.7090                     & 0.7610                     & 0.8210                     & 0.7580                     & 0.7670                     & 0.8830                     & 0.7450                     & 0.8350                     & 0.8280                     & 0.7310                     & 0.8460                      & 0.8066                      \\
\multicolumn{1}{c|}{JT-100 \cite{lenga2020continual}}             & \multicolumn{1}{l}{0.9211} & \multicolumn{1}{l}{\textbf{0.8370}} & \multicolumn{1}{l}{0.9179} & \multicolumn{1}{l}{0.7005} & \multicolumn{1}{l}{0.7940} & \multicolumn{1}{l}{0.8355} & \multicolumn{1}{l}{0.7651} & \multicolumn{1}{l}{0.7842} & \multicolumn{1}{l}{0.8685} & \multicolumn{1}{l}{0.7572} & \multicolumn{1}{l}{0.8506} & \multicolumn{1}{l}{0.8343} & \multicolumn{1}{l}{0.7142} & \multicolumn{1}{l|}{0.8843} & \multicolumn{1}{l}{0.8189}  \\
\hline\hline
Ours                                & \textbf{0.9421}            & 0.8234            & \textbf{0.9389}            & \textbf{0.7150}            & \textbf{0.8013}            & \textbf{0.8394}            & \textbf{0.7878}            & \textbf{0.7884}            & \textbf{0.9057}            & \textbf{0.7607}            & \textbf{0.8580}            & \textbf{0.8403}            & \textbf{0.7466}            & \textbf{0.9000}             & \textbf{0.8320}             \\
\hline
\end{tabular}}
\end{table*}

\begin{table*}[ht]
\caption{Comparison with State-of-the-arts on NIH Dataset}\label{SOTA Comparison Table}
\centering
\resizebox{\textwidth}{!}{\begin{tabular}{c|cccccccccccccc|c} 
\hline
\multirow{2}{*}{Method} & \multicolumn{7}{c|}{NIH Only}                                                                                                               & \multicolumn{7}{c|}{Common}                                                                                                                 & \multirow{2}{*}{Mean}  \\ 
\cline{2-15}
                        & Emph             & Fibr             & Hern           & Infi             & P\_T             & Mass             & \multicolumn{1}{c|}{Nodu} & Atel             & Card             & Cons             & Edem             & Effu             & Pne1             & \multicolumn{1}{c|}{Pne2} &                        \\ 
\hline
\hline
DCNN \cite{wang2017chestx}                    & 0.8150           & 0.7690           & 0.7670           & 0.6090           & 0.7080           & 0.7060           & 0.6710                    & 0.7160           & 0.8070           & 0.7080           & 0.8350           & 0.7840           & 0.6330           & 0.8060           & 0.7381                 \\
LSTM-Net \cite{yao2017learning}                & 0.8420           & 0.7570           & 0.8240           & 0.6750           & 0.7240           & 0.7270           & 0.7780                    & 0.7330           & 0.8580           & 0.7170           & 0.8060           & 0.8060           & 0.6900           & 0.8050           & 0.7673                 \\
TieNet \cite{wang2018tienet}                  & 0.8650           & 0.7960           & 0.8760           & 0.6660           & 0.7350           & 0.7250           & 0.6850                    & 0.7320           & 0.8440           & 0.7010           & 0.8290           & 0.7930           & 0.7200           & 0.8470           & 0.7724                 \\
AGCL \cite{tang2018attention}                    & 0.9075           & 0.8179           & 0.8747           & 0.6892           & 0.7647           & 0.8136           & 0.7545                    & 0.7557           & 0.8865           & 0.7283           & 0.8475           & 0.8191           & 0.7292           & 0.8499           & 0.8027                 \\
CRAL \cite{guan2018multi}                    & 0.9080           & 0.8300           & 0.9170           & 0.7020           & 0.7780           & 0.8340           & 0.7730                    & 0.7810           & 0.8800           & 0.7540           & 0.8500           & 0.8290           & 0.7290           & 0.8570           & 0.8159                 \\
CheXNet \cite{rajpurkar2017chexnet}                 & 0.9249           & 0.8219           & 0.9323           & 0.6894           & 0.7925           & 0.8307           & 0.7814                    & 0.7795           & 0.8816           & 0.7542           & 0.8496           & 0.8268           & 0.7354           & 0.8513           & 0.8180                 \\
LLAGNet \cite{chen2019lesion}                 & 0.9390           & 0.8320           & 0.9160           & 0.7030           & 0.7980           & 0.8410           & 0.7900                    & 0.7830           & 0.8850           & 0.7540           & 0.8510           & 0.8340           & 0.7290           & 0.8770           & 0.8237                 \\
Yan et al. \cite{yan2018weakly}              & \textbf{0.9422}  & 0.8326           & 0.9341           & 0.7095           & \textbf{0.8083}  & \textbf{0.8470}  & \textbf{0.8105}           & \textbf{0.7924}  & 0.8814           & 0.7598           & 0.8470           & 0.8415           & 0.7397           & 0.8759           & 0.8302                 \\ 
\hline
\hline
Ours w/ TenCrop         & 0.9396           & \textbf{0.8381}           & \textbf{0.9371}  & \textbf{0.7184}  & 0.8036           & 0.8376           & 0.7985                    & 0.7891           & \textbf{0.9069}  & \textbf{0.7681}  & \textbf{0.8610}  & \textbf{0.8418}  & \textbf{0.7419}           & \textbf{0.9063}           & \textbf{0.8349}        \\
\hline
\end{tabular}}
\end{table*}

\begin{figure*}[ht]
\centering
\includegraphics[width=\textwidth]{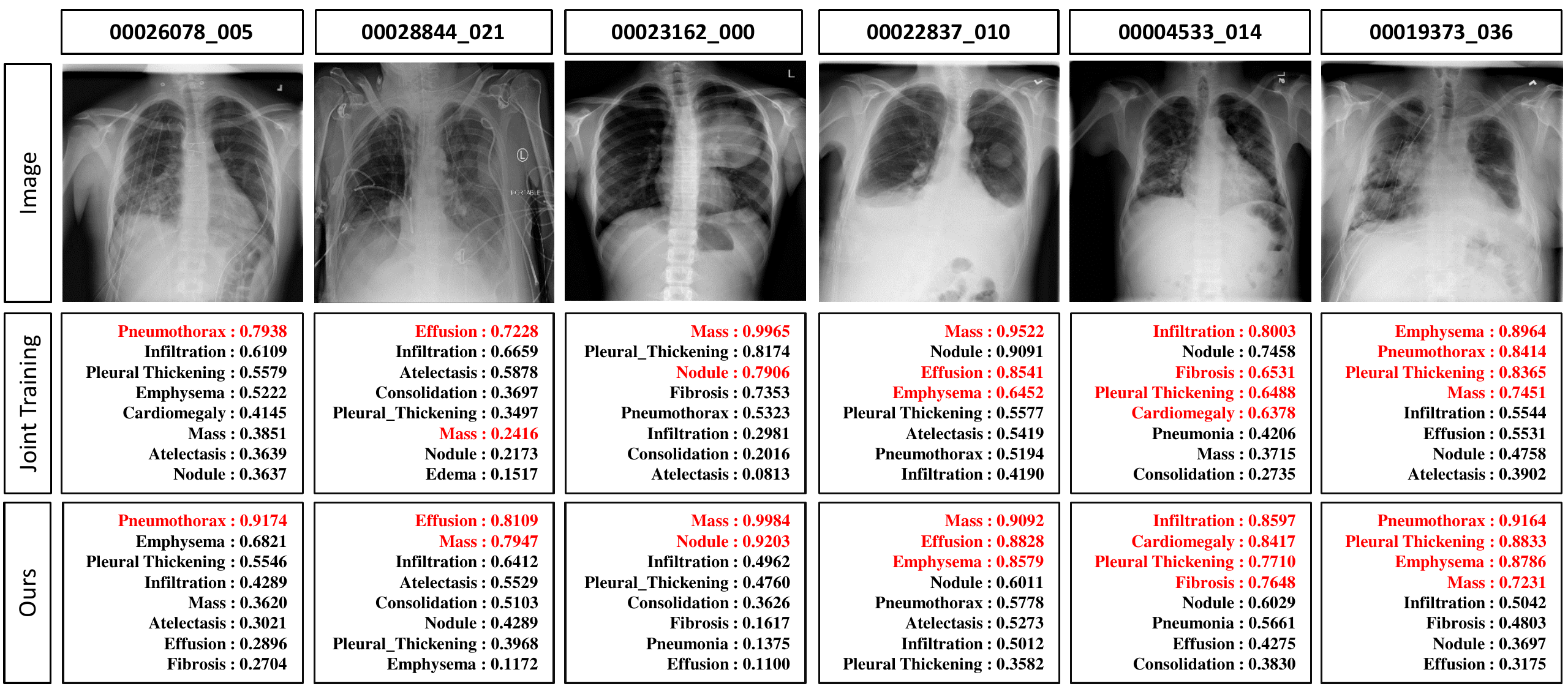}
\caption{Top-8 predicted findings and the corresponding prediction scores of joint trained DenseIBN-121 and our method. The ground truth labels are highlighted in \textcolor{red}{Red}. The number on the top of the image is its corresponding file name. Best viewed in color.} 
\label{Sample Prediction}
\end{figure*}

\begin{table*}
\centering
\caption{Classification Results On ImsightCXR Dataset Incorporating NIH as External Dataset}
\label{ImsightCXR Comparison Table}
\resizebox{\textwidth}{!}{\begin{tabular}{c|cccccccccccccc|c} 
\hline
\multirow{2}{*}{Experiments} & \multicolumn{2}{c|}{ImsightCXR Only}                   & \multicolumn{12}{c|}{Common}                                                                                                                                                                                               & \multirow{2}{*}{Mean}  \\ 
\cline{2-15}
                             & Fracture        & \multicolumn{1}{c|}{Tuberculosis} & Atel            & Card            & Cons            & Fibr            & Hern            & Infi            & Mass            & Nodu            & Effu            & P\_T & Pne1            & Pne2            &                        \\ 
\hline
\hline
Single Data                  & 0.916           & 0.909                             & 0.888           & 0.966           & 0.884           & 0.802           & 0.910           & \textbf{0.826}           & 0.891           & 0.807          & 0.950           & 0.819                & 0.860           & 0.955           & 0.884                  \\
Joint Training               & \textbf{0.922}           & 0.916                             & 0.897           & \textbf{0.976}  & 0.885           & 0.814           & 0.948           & 0.768           & 0.916           & 0.838           & 0.956           & 0.844                & \textbf{0.870}  & \textbf{0.964}  & 0.894                  \\ 
\hline
\hline
Ours                         & \textbf{0.922}  & \textbf{0.920}                    & \textbf{0.898}  & 0.975           & \textbf{0.897}  & \textbf{0.820}  & \textbf{0.968}  & 0.815  & \textbf{0.923}  & \textbf{0.840}  & \textbf{0.957}  & \textbf{0.845}       & 0.865           & 0.962           & \textbf{0.901}         \\
\hline
\end{tabular}}
\end{table*}

\begin{table*}[]
\centering
\caption{Ablation Study on Different Components of Our Method. Mean$_{Com}$: the average AUC of common categories shared across datasets; Mean$_{Int}$: the average AUC of categories labeled in NIH only.}
\label{Ablation Study Table}
\begin{tabular}{|l|c|c|c|c|c|c|c|c|c|}
\hline
Indices                                                    & a                         & b                         & c                                & d                                & e                                & \textcolor{black}{f}    & g                                & h                         & i                                \\ \hline
Joint                                                      & \checkmark & \checkmark & \checkmark        & \checkmark        & \checkmark        & \textcolor{black}{\checkmark} & \checkmark        & \checkmark & \checkmark        \\
TW                                                         &                           & \checkmark & \checkmark        & \checkmark        & \checkmark        &                           & \checkmark        & \checkmark & \checkmark        \\
TAT                                                        &                           &                           & \checkmark        & \checkmark        &                                  & \checkmark & \textcolor{black}{\checkmark}        &                           & \checkmark        \\
TE                                                         &                           &                           &                                  & \checkmark        & \checkmark        & \textcolor{black}{\checkmark} & \checkmark        & \checkmark &                                  \\
Uncertain                                                  &                           &                           &                                  &                                  & \checkmark        & \textcolor{black}{\checkmark} & \checkmark        & \checkmark &                                  \\
HAT                                                        &                           &                           &                                  &                                  &                                  &                           &                                  & \checkmark &                                  \\
Hard Label                                                 &                           &                           &                                  &                                  &                                  &                           &                                  &                           & \checkmark        \\ \hline
Emph                                                       & 0.9368                    & 0.9389                    & 0.9414                           & 0.9410                           & 0.9409                           & \textcolor{black}{0.9368}                    & \textbf{0.9421} & 0.9373                    & 0.9361                           \\
Fibr                                                       & 0.8199                    & 0.8206                    & \textbf{0.8295} & 0.8257                           & 0.8242                           & \textcolor{black}{0.8233}                    & 0.8234                           & 0.8215                    & 0.8101                           \\
Hern                                                       & 0.9349                    & 0.9267                    & 0.9388                           & 0.9376                           & 0.9300                           & \textcolor{black}{0.9445}                   & 0.9389                           & 0.9193                    & \textbf{0.9495} \\
Infi                                                       & 0.7138                    & 0.7083                    & 0.7107                           & 0.7155                           & 0.7121                           & \textcolor{black}{0.7108}                    & \textbf{0.7150} & 0.7126                    & 0.7051                           \\
P\_T                                                       & 0.7918                    & 0.7956                    & 0.7999                           & 0.7906                           & 0.8002                           & \textcolor{black}{0.7960}                    & \textbf{0.8013} & 0.7990                    & 0.7919                           \\
Mass                                                       & 0.8205                    & 0.8251                    & 0.8308                           & 0.8217                           & 0.8343                           & \textcolor{black}{0.8296}                    & \textbf{0.8394} & 0.8227                    & 0.8353                           \\
Nodu                                                       & 0.7725                    & 0.7852                    & 0.7865                           & 0.7759                           & 0.7827                           & \textcolor{black}{0.7807}                    & \textbf{0.7878} & 0.7771                    & 0.7801                           \\
Atel                                                       & 0.7590                    & 0.7863                    & 0.7862                           & 0.7864                           & 0.7876                           & \textcolor{black}{0.7651}                    & \textbf{0.7884} & 0.7808                    & 0.7756                           \\
Card                                                       & 0.9024                    & 0.9086                    & 0.9052                           & 0.9095                           & \textbf{0.9086} & \textcolor{black}{0.8982}                    & 0.9057                           & 0.8990                    & 0.9046                           \\
Cons                                                       & 0.7486                    & 0.7579                    & 0.7636                           & 0.7584                           & \textbf{0.7639} & \textcolor{black}{0.7481}                    & 0.7607                           & 0.7517                    & 0.7537                           \\
Edem                                                       & 0.8296                    & 0.8532                    & 0.8571                           & 0.8553                           & \textbf{0.8606} & \textcolor{black}{0.8370}                    & 0.8580                           & 0.8396                    & 0.8485                           \\
Effu                                                       & 0.8279                    & 0.8396                    & \textbf{0.8422} & 0.8405                           & 0.8404                           & \textcolor{black}{0.8308}                    & 0.8403                           & 0.8344                    & 0.8360                           \\
Pne1                                                       & 0.7352                    & 0.7354                    & 0.7367                           & 0.7442                           & 0.7398                           & \textcolor{black}{0.7354}                    & \textbf{0.7466} & 0.7371                    & 0.7342                           \\
Pne2                                                       & 0.8870                    & 0.8944                    & 0.8998                           & \textbf{0.9045} & 0.8986                           & \textcolor{black}{0.8814}                    & 0.9000                           & 0.8957                    & 0.8911                           \\ \hline
\textcolor{black}{Mean$_{Com}$} & 0.8128                    & 0.8251                    & 0.8273                           & 0.8284                           & \textbf{0.8285} & \textcolor{black}{0.8137}                    & \textbf{0.8285} & 0.8198                    & 0.8205                           \\ \hline
\textcolor{black}{Mean$_{Int}$} & 0.8272                    & 0.8286                    & 0.8339                           & 0.8297                           & 0.8321                           & \textcolor{black}{0.8317}                    & \textbf{0.8354} & 0.8271                    & 0.8297                           \\ \hline
Mean                                                       & 0.8200                    & 0.8268                    & 0.8306                           & 0.8291                           & 0.8303                           & \textcolor{black}{0.8227}                    & \textbf{0.8320} & 0.8234                    & 0.8251                           \\ \hline
\end{tabular}
\end{table*}

\subsection{Classification Results on NIH Dataset}\label{Classification Results on NIH Incorporating External Dataset}

\subsubsection{NIH incorporating CheXpert as external dataset}\label{Approaches Incorporating External Dataset}
We first conduct experiments with NIH being internal and another large dataset being external to demonstrate the knowledge mining capability of our method. We report in Table \ref{External Domain Comparison Table} the comparison among several other methods and some variants of our method: a) a deep network trained on NIH only (Single Dataset); b) a basic partial-label classification network that merges the labels of NIH and CheXpert (Joint Training); c) a multi-task network which takes CheXpert as the external domain and treats same labels from two domains as independent two categories following \cite{guendel2018learning}; d) a location-aware Dense Network (DNetLoc) \cite{guendel2018learning} which utilizes extra 297,541 chest X-ray images from PLCO dataset \cite{PLCO}; e). a deep model jointly trained (JT-100 \cite{lenga2020continual}) with full NIH and additional 180,000 images from MIMIC-CXR \cite{MIMIC-CXR}. a), b) and c) are based on DenseIBN-121 backbone, while d) and e) are based on Dense-121 backbone.\\
\indent As can be observed from Table \ref{External Domain Comparison Table}, four out of five methods incorporating an external domain exceed the performance of Single Dataset. The multi-task-based models (DNetLoc and Multi-Task) show lower AUC than the joint training methods (Joint Training and JT-100), which implies the information shared between two datasets may not be fully utilized when separating the same label from different domains to different categories. Meanwhile, both joint training based models achieve AUC about $0.8200$, demonstrating that merging the classes from multiple domains could benefit to model and help mine the common knowledge shared across datasets. Nevertheless, simply merging the classes and joint training a model is not enough for leveraging the external domain, as the domain discrepancy and label discrepancy have not been addressed. Our method reaches the best performance with a large margin with $1.2$ AUC gain compared with others, which demonstrates that solving the dataset discrepancies help mine the common knowledge and hidden knowledge more effectively. We also show in Fig. \ref{Sample Prediction} with typical classification results of the joint training model and our method for a qualitative comparison.

\subsubsection{Comparison with state-of-the-arts on NIH}

We further compare our method with other state-of-the-art approaches. We choose a number of typical existing models including: deep convolutional neural networks (DCNN) \cite{wang2017chestx}, deep network with long-short-term-memory unit (LSTM-Net) \cite{yao2017learning}, Text-Image Embedding network (TieNet) \cite{wang2018tienet}, Attention-Guided Curriculum Learning (AGCL) \cite{tang2018attention}, Category-wise Residual Attention Learning (CRAL) \cite{guan2018multi}, CheXNet \cite{rajpurkar2017chexnet}, Lesion location Attention Guided Network (LLAGNet) \cite{chen2019lesion}, and the method of Yan et al. \cite{yan2018weakly}. All these works adopt the official data split except for LSTM-Net. For CheXNet, we report the results from the implementation of Yan et al., which followed the same official split. \\
\indent For a fairer comparison, we follow \cite{yan2018weakly} and take advantage of horizontal flipping and random crop to augment training data and using average probabilities of ten cropped sub-images (four corner crops and one central crop plus horizontally flipper version of these) as the final prediction. As shown in Table \ref{SOTA Comparison Table}, the method of Yan et al. is the previous state-of-the-art with AUC $0.8302$, while our method exceeds all the compared models and achieves a new state-of-the-art performance of $0.8349$ AUC. Specifically, our classification results outperform others in 9 out of 14 categories. In more detail, our method surpasses all others in 6 out of 7 common diseases shared between NIH dataset and CheXpert, while mining the unlabeled data also helps boost the performance to superior or comparable performance on the classes labeled in NIH only. Overall, our method achieves $0.47$  average AUC improvement than the former best method by mining the knowledge from the external dataset. We point out that other joint training methods in Table \ref{External Domain Comparison Table} do not achieve comparable results with the state-of-the-art. Therefore, the comparison in Table \ref{SOTA Comparison Table} is meaningful. It is also worth noting that, we designed easy-to-implement losses to mine the knowledge from the external dataset, instead of designing complex models or ensembling multiple models.

\subsection{Classification Results on ImsightCXR Dataset}\label{Experiments on ImsightCXR Dataset}

To demonstrate the general effectiveness of our method, we carry out experiments on the private dataset ImsightCXR with NIH as an external domain. ImsightCXR shares 12 common pathology labels with NIH, while it contains two additional pathological findings: fracture and tuberculosis. As there are no other works for comparison, we implement three methods: the Single Dataset model, the Joint Training model, which is the best strategy other than ours in Section \ref{Approaches Incorporating External Dataset}, and our own method. We report in Table \ref{ImsightCXR Comparison Table} the quantitative classification results. As could be observed, incorporating external domain benefits to the performance on ImsightCXR with at least $1$ point gain in the average AUC (from $0.884$ for Single Dataset to $0.894$ for Joint Training). By encouraging the model to mitigate the domain variance and learn from unknown labels, our network further achieves $0.901$ AUC with $0.7$ performance gain compared with Joint Training. Specifically, our approach shows supreme performance on 10 out of 14 categories and comparable to the best results on the rest four classes. Also, the AUC increase on the common labels contributes more to the overall performance gain, which indicates the additional information of the common categories is more sufficient.

\subsection{Analysis of Different Components}\label{Ablation study on Different Components}
\subsubsection{Ablation study on each method}

We report in Table \ref{Ablation Study Table} the effect of each component of our model. 
In addition to task weighting (TW), task-specific adversarial training (TAT), temporal ensembling (TE), and uncertainty mechanism, we also compare with our variants: ours replacing TAT with holistic adversarial training (HAT), where the network ends with an $N \times C$ FC-layer and generates class scores based on the single holistic feature; and ours replacing UTE with Hard Label assignment by model ensembles. Checkmarks in the table indicate the specific component is incorporated. We compute the AUC of every disease, the mean AUC of all (see "Mean" in the table). To see the influence of the specific design on different categories, we as well report the mean AUC of the seven diseases labeled in NIH dataset only ("Mean$_{Com}$") and the mean AUC of the seven diseases shared across two domains("Mean$_{Int}$").\\
\indent Although adding the external dataset, i.e., CheXpert, has already enriched a large number of positive samples to NIH, the AUC of Joint Training is limited to $0.8200$. This indicates the knowledge from the external domain could be more effectively mined and utilized. Meanwhile, adopting TW strategy leads to $0.68$ improvement, which demonstrates our weighted task formulation suits the setting of our problem (see experiment b). Specifically, the major profits brought by TW are from the NIH-only-classes, where "Mean$_{Com}$" is increased with over 1 point. Based on TW, adding TAT helps the model reach an average AUC of $0.8306$, and gain about $0.6$ average points on the common labels (see "Mean$_{Int}$" in experiment c). Further, we find directly incorporating temporal ensembling without uncertainty would lead to a performance drop (from $0.8320$ to $0.8291$; see experiments g and d), while incorporating uncertainty-aware temporal ensembling (UTE) even without TAT would raise the average AUC to $0.8302$ (see experiments e). In particular, TAT reaches the second-highest "Mean$_{Int}$" performance ($0.8339$), and UTE achieves the second-highest "Mean$_{Com}$" performance ($0.8285$) among all. We also observe a severe performance degradation (AUC drops to $0.8227$ in experiment f) when removing TW from our model. This result, combined with the finding in experiment b, indicate TW plays an essential role in our proposed model.
Despite that experiments c and e have shown comparable performance with previous state-of-the-art \cite{yan2018weakly}, our method could only reach its best performance when unifying all proposed components together (see experiment g). This implies that TAT and UTE at a certain level are complementary to each other. Our model shows a $1.2$ performance gain in average AUC compared with simple joint training. Moreover, it reaches the best "Mean$_{Com}$" and "Mean$_{Int}$" performance among all experiments, which implies that the proposed components could collectively benefit the model.\\
\indent We additionally investigate the effectiveness of TAT and UTE by replacing either one with another component. It can be observed in experiment h that by changing TAT to HAT, the average AUC for all categories dropped drastically to $0.8234$. As argued before, the single holistic feature vector would carry too rich semantics. Since the label distribution varies across CXR datasets, the discriminator could easily distinguish the source domains. Therefore, adversarial training could dominate the classification task and lead to poor performance. We also replace UTE with hard pseudo-label methods in experiment i. As a result, the overall performance decreases with nearly $0.7$ points (from $0.8320$ to $0.8251$), which demonstrates the advantage of leveraging soft labels as the target for unknown labels. 

\subsubsection{Experiments on the hyperparameters}
\begin{figure}[t]
\centering
{
\begin{minipage}[t]{\linewidth}
 \centering
 \subfigure[Change of AUC on NIH with different $\lambda_{ute}$ when $\lambda_{tat}$ is fixed to $0.03$. Red curve: overall mean AUC; Green curve: mean AUC over NIH-only-categories.]{
\includegraphics[width=0.75\linewidth, height=5cm]{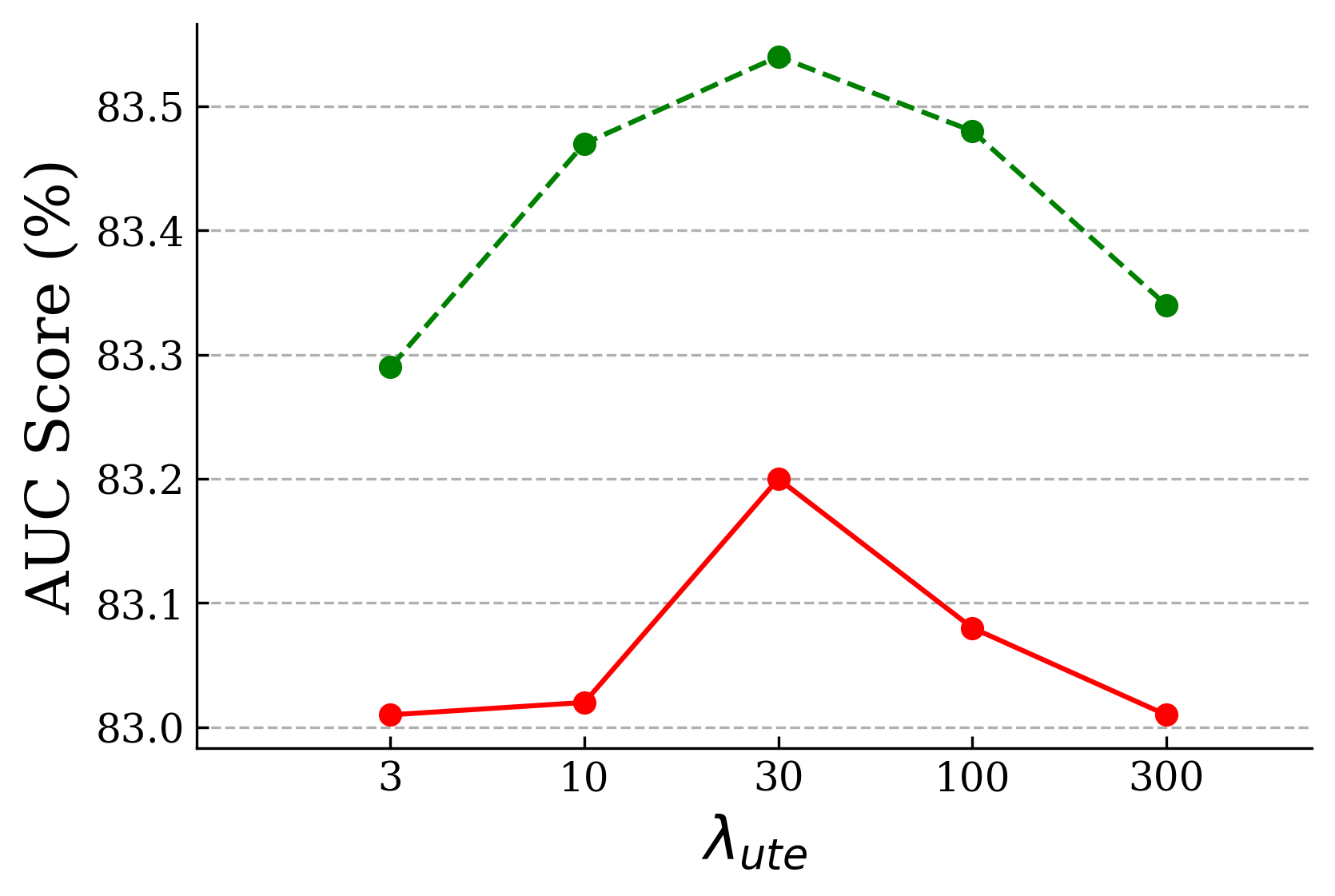}
\label{hyperparameter-TE}}
\end{minipage}
}
{
\begin{minipage}[t]{\linewidth}
\centering
\subfigure[Change of AUC on NIH with different $\lambda_{tat}$ when $\lambda_{ute}$ is fixed to $30$. Red curve: overall mean AUC; blue curve: mean AUC over common categories.]{
\includegraphics[width=0.75\linewidth, height=5cm]{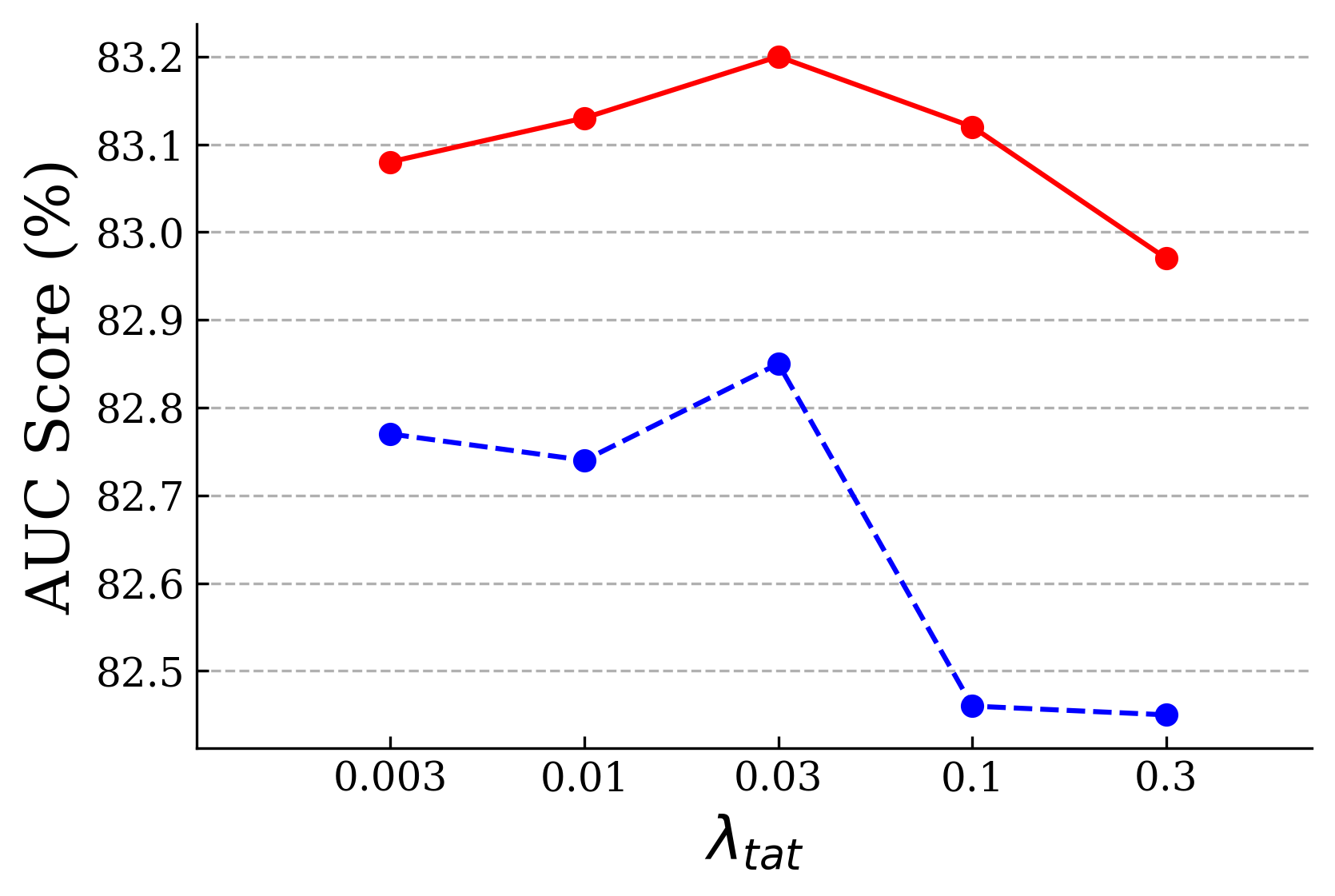} 
\label{hyperparameter-DC}}
\end{minipage}
}
\centering
    \caption{Diseases classification average AUC scores on NIH dataset with different values of hyperparameters $\lambda_{tat}$ and $\lambda_{ute}$. \textcolor{red}{Red} curve represents the mean AUC of all 14 categories, \textcolor{blue}{blue} curve stands for the average performance of common labels, and \textcolor{green}{green} curve means the average score of NIH only classes. Best viewed in color.}
\label{hyperparameter}
\end{figure}

We further analyse the influence of the hyper-parameters $\lambda_{tat}$ and $\lambda_{ute}$ with the experiment on NIH dataset. Based on the model with task weights, we first fix $\lambda_{tat}$ to be $0.03$ and vary $\lambda_{ute}$ from 3 to 300 with a factor of approximate 3 for each model. As shown in Fig. \ref{hyperparameter-TE}, both mean AUC and Mean$_{Int}$ are consistently increasing when rising the of $\lambda_{ute}$ from $3$ to $30$ and consistently decreasing when further rising $\lambda_{ute}$. We also observe that mean AUC is larger than $0.8300$ under all the settings for $\lambda_{ute}$. We then fix the value of $\lambda_{ute}$ to 30 and vary $\lambda_{tat}$ from $0.003$ to $0.3$ with a factor of approximate 3 for each model. As shown in Fig. \ref{hyperparameter-TE}, the performance of the model is stable when varying $\lambda_{tat}$ from $0$ to $0.1$. In addition, we could see a big drop in the average AUC score for the common classes when $\lambda_{tat}$ set to $0.3$, which implies the adversarial loss dominates the training of the network. Both mean AUC and Mean$_{Com}$ reach the peaks when $\lambda_{tat}=0.03$.

\section{Discussion}
This work tackles the bottleneck of joint training on imperfect chest radiograph datasets. The obstacle arises from two aspects: domain discrepancy and label discrepancy. Previously, several works have been done on identifying the domain shift in the sense of different image distributions \cite{PLOSCrossSection,cohen2020limits}, while few studies how to solve this problem in joint training. Meanwhile, some studies have been carried out on training a deep learning model with CXR images from multiple domains. Yao et al. \cite{yao2019strong} attempted to develop a binary classification model to identify radiographs with pathological findings from those with no finding. However, the definition of "no finding" is inconsistent among different sites, which might be one of the main reasons that limited improvements are observed. Zhang et al. \cite{zhang2019improve} presented domain generalization approaches that leverage training on several datasets to improve the model's performance on an unseen target set. Their approaches mainly focus on learning more general features, while neglecting the label discrepancy. Lenga et al. \cite{lenga2020continual} consider a continual learning problem in which the model is required to "remember" the learned knowledge when trained on a new domain. They conduct experiments on different strategies and find out that joint training achieves the best AUC score compared with other learning-without-forget approaches. Nevertheless, none of these studies have addressed the performance bottleneck for joint training. \\
\indent The goal of Guendel et al. \cite{guendel2018learning} is the closest to ours, as they also aim at mining the location knowledge from the PLCO dataset to improve NIH performance. They consider the possibility that the datasets might be created based on different label definitions and propose a multi-task network that treats the labels from different datasets independently. Cohen et al. \cite{cohen2020limits} also point out the label inconsistency limits the generalization performance of deep models. However, our experiment on the multi-task setting shows little improvement from the model trained on a single dataset. This finding implies that though some level of label inconsistency exists in between datasets, a larger part of the label definitions ought to be similar. Hence, the multi-task model could fail to integrate the knowledge across domains. Up till now, the most robust way to deal with label inconsistency might still need the participation of radiologists. For instance, one could label different datasets with a more consistent and rigorous reference standard \cite{majkowska2020chest}. One may also consider leveraging the inter-observer uncertainty when annotating the dataset \cite{peterson2019human} to train more robust models. How to take advantage of deep learning methods to address this problem is still an open question.\\
\indent Apart from the label inconsistency and label discrepancy we explored, there are also discussions about the CXR label noises. The annotations of public CXR datasets are often extracted by natural language processing from radiological reports, which inevitably results in a certain level of label noise. One common way to mitigate the label noise is to hand-label the images by radiologists as we have done for ImsightCXR. Recently, Majkowska et al. \cite{majkowska2020chest} have relabeled a subset (1,962 images) of NIH testing data with at least three radiologists per image. We then further test on this relabeled set to see whether our model is also robust to the noise. To be consistent with the annotations used in the paper, we average the outputs of Edema, Consolidation, Mass, Nodule, Pneumonia, Atelectasis, and Infiltration to be the prediction of "Airspace Opacity". We also average the outputs of Mass and Nodule to be the prediction of "Mass or Nodule". The methods show consistent improvements in Table \ref{Classification Results On NIH Clean Subset}. With TenCrop, the proposed method further achieves competitive AUC with the ensemble model in \cite{majkowska2020chest} on "Pneumothorax" ($0.925$ v.s. $0.94$). Note that the compared model \cite{majkowska2020chest} was trained on a less noisy large dataset, from which radiologists relabeled 37,521 radiographs. Overall, this experiment demonstrates the robustness of the proposed method to label noise.\\
\begin{table}[t]
\color{black}
\centering
\caption{Classification Results On NIH Clean Subset}
\label{Classification Results On NIH Clean Subset}
\resizebox{0.48\textwidth}{!}{\begin{tabular}{l|ccc} 
\hline
Method          & Pneumothorax    & Airspace Opacity & Nodule or Mass   \\ 
\hline \hline
Single Data          & 0.870           & 0.899            & 0.843            \\ 
\hline
Joint Training          & 0.899           & 0.901            & 0.859            \\ 
\hline
Ours            & \textbf{0.913}  & \textbf{0.902}   & \textbf{0.860}   \\ 
\hline \hline
Ours w/ TenCrop & 0.925           & 0.903            & 0.874            \\
\hline
\end{tabular}}
\end{table}
\indent One limitation of our present work is that we focus on adopting a large-scale external domain to improve the internal one's performance. As aforementioned, one reason the external dataset could benefit the internal during joint training is that incorporating the external dataset brings a lot more positive examples. We also tried to incorporate two datasets, i.e., CheXpert and MIMIC-CXR, as the external set to improve NIH. The mean AUC on NIH with TenCrop is $0.8353$. This result implies that there might be a saturated region of adding more external data for training when the extra samples do not bring many times more positive samples. Moreover, it might be more scalable training with three or more datasets and learning a general model that can perform well on every set. Clinically, the automatic diagnosis assistant tool is also required to cover more diseases for general needs. Some studies began to on this complex scenario, such as universal representation learning \cite{rebuffi2017learning,Liu2020MSNet}. Our future work would refer to these works and consider more complex neural network design to enable a universal model. Moreover, the TW, TAT, and UTE modules are easily plug-and-play and could be improved separately. The future work would also figure out how to improve our method based on current partial-label formulation. For example, the task weights could be calculated by comparing the gradients of different tasks \cite{chen2018gradnorm}, the adversarial training in the TAT module could be implemented with more advanced GANs \cite{gulrajani2017improved}, and the UTE module may draw advantage from unifying other semi-supervised learning mechanisms \cite{berthelot2019mixmatch,wang2020towards}. \\
\indent In general, how to develop automatic thoracic disease screening systems that are general on many domains and incorporate more and more imperfect data is still an open question. There are many other research topics concerning training with multiple medical image datasets, such as domain generalization \cite{zhang2019improve,zhang2020generalizing}, incremental learning \cite{lenga2020continual}, universal representation learning \cite{Liu2020MSNet}, etc., and coping with imperfect external data is an unavoidable problem for all. Beyond learning with multi-site, there are more questions to be answered for CXR diseases recognition, such as how to utilize the label relations \cite{yan2019holistic} instead of modeling multi-label classification as independent binary tasks, how to tackle the noisy label problem as many large-scale datasets are labeled by NLP tools, or how to take into consideration the inter-observer and inter-dataset annotation variance when labeling the data. It is also worth studying what factors (age, patient cohort, scanner, etc.) cause the most domain distribution shift if we could access the patient demographic from multiple datasets. Our future work would dig further to explore these problems for joint training on multiple Chest X-ray datasets.

\section{Conclusion}
In this paper, we alleviate the bottleneck of training with imperfect chest X-ray data when incorporating an external dataset. We point out the obstacles that exist in both domain discrepancy and label discrepancy. We formulate the arisen partial-label classification problem into weighted independent binary tasks for further coping with the challenges. To address the domain discrepancy, we present task-specific adversarial training to learn more domain-invariant features specifically for the common categories. To address the label discrepancy, we present uncertainty-aware temporal ensembling to gradually and reliably mine the hidden knowledge from unknown categories. To the best of our knowledge, this is the first work that simultaneously addresses domain and label discrepancies for integrating large-scale medical image datasets. We conduct extensive experiments on both public datasets and a private dataset and achieve state-of-the-art performance on NIH dataset by mining knowledge from external data, demonstrating the effectiveness of our method. Our framework is general for training deep neural networks with multiple sites and could be extended to further practical clinical usages.

\bibliographystyle{IEEEtran}
\bibliography{ref}
\end{document}